\documentclass[pdflatex,sn-mathphys-num]{sn-jnl}

\usepackage{amsmath,amssymb,amsfonts}
\usepackage{amsthm}
\usepackage{mathrsfs}

\usepackage{graphicx}
\usepackage{xcolor}
\usepackage{listings}

\usepackage{booktabs}
\usepackage{multirow}
\usepackage{array}
\usepackage{tabularx}
\usepackage{caption}
\usepackage{adjustbox}

\usepackage[title]{appendix}

\usepackage{manyfoot}

\usepackage{algorithm}
\usepackage{algorithmicx}
\usepackage{algpseudocode}

\usepackage{float}

\usepackage[T1]{fontenc}
\usepackage[utf8]{inputenc} 
\usepackage{lmodern}
\usepackage{textcomp}
\usepackage{newunicodechar}
\newunicodechar{–}{--}
\newunicodechar{—}{---}
\newunicodechar{’}{'}
\newunicodechar{“}{``}
\newunicodechar{”}{''}

\usepackage{listings}

\usepackage{xurl} 

\theoremstyle{thmstyleone}%
\theoremstyle{thmstyletwo}%

\theoremstyle{thmstylethree}%

\usepackage{parskip} 

\begin{document}

\title[Self-Reflective Chain-of-Thought (SRCoT): Evidence-Based Text Generation via a Closed-Loop Retrieve–Reflect–Revise Pipeline on Edge LLMs]{Self-Reflective Chain-of-Thought (SRCoT): Evidence-Based Text Generation via a Closed-Loop Retrieve–Reflect–Revise Pipeline on Edge LLMs}

\author*[1]{\fnm{Fabian} \sur{Broszio}}\email{fabian.broszio@fu-berlin.de}

\author*[1]{\fnm{Michael} \sur{Hoffmann}}\email{michaeh78@zedat.fu-berlin.de}

\author*[1]{\fnm{Adrian} \sur{Paschke}}\email{adrian.paschke@fu-berlin.de}

\affil[1]{\orgdiv{Department of Mathematics and Computer Science}, \orgname{Freie Universität Berlin}, \orgaddress{\street{Arnimallee 14}, \city{Berlin}, \postcode{14195}, \country{Germany}}}

\abstract{Large language models (LLMs) can generate fluent outputs that lack source support or conflict with local operational constraints. We introduce Self-Reflective Chain-of-Thought (SRCoT), a modular architecture that organizes generation as a closed Retrieve--Reflect--Revise loop. SRCoT separates environment state, persistent knowledge, accumulated experience, and short-term memory into provenance-labeled stores. BM25 retrieval supplies local context; subsequent passes expose intermediate drafts, missing-information markers, follow-up queries, and optional claim--evidence mappings. These artifacts inform retrieval, revision, clarification, or abstention decisions. We examine the architecture through qualitative, trace-based case studies in two fully synthetic settings: emergency medical services and Munich-Sim, an emergency-operations simulation. Local prototype runs on an Apple M4-family laptop with 128\,GB memory illustrate source-labeled documentation, explicit gap marking, and structured dispatch-style outputs. The traces also reveal provenance-assignment errors, incomplete evidence coverage, and unsafe assumptions that remain possible despite prompting constraints. These cases demonstrate report-level inspectability and local execution, not comparative superiority, hallucination reduction, clinical validity, or operational safety. The architecture separates retrieval and control from the model interface, but behavior across different models remains to be evaluated. SRCoT is a research proof of concept requiring human review, with its principal contribution lying in an inspectable control procedure rather than a new model.}


\maketitle

\section{Introduction}
\label{Introduction}
Large Language Models (LLMs) have achieved strong performance across a wide range of natural language processing tasks, including text generation, summarization, and interactive reasoning. Their fluency and apparent coherence have motivated increasing interest in deploying them in applied and decision-support settings. However, despite these advances, LLMs remain prone to generating outputs that are convincing in form but unsupported by evidence, internally inconsistent, or misaligned with domain-specific constraints. Such failures—often referred to as hallucinations\footnote{In LLMs, ``hallucination'' refers to fluent output that is not supported by the provided sources or is fabricated/inaccurate, even if it sounds plausible.}—are particularly problematic in safety-critical or operational environments, where erroneous outputs may be difficult to detect yet carry substantial consequences \cite{ji_hallucination_survey_2023,islam2024hallucination_mitigation_survey,maynez2020faithfulness}.

A contributing factor to this problem is the dominant usage paradigm of current LLM systems. Most deployments rely on a single, monolithic context in which persistent knowledge, task-specific state, intermediate reasoning, and action decisions are entangled within a shared prompt or hidden representation \cite{yao_react_2022, packer_memgpt_2023}. While recent approaches such as chain-of-thought prompting \cite{wei2022chain}, retrieval-augmented generation (RAG)\footnote{Retrieval-augmented generation (RAG) pairs a Large Language Model (LLM) with a retrieval step over an external corpus: relevant passages are fetched for a query and inserted into the model’s context so that generation can be grounded in retrieved evidence. This improves traceability (sources can be cited) but remains limited by retrieval quality and context use.} \cite{lewis2020retrieval}, and tool use improve reasoning performance, they typically lack an explicit separation of different knowledge and state types, making it difficult to monitor, verify, or control the generation process in a principled manner \cite{packer_memgpt_2023, yao_react_2022}.

In this work, we argue that reliable and evidence-based text generation requires a shift toward structured, closed-loop reasoning architectures. We draw inspiration from early cybernetics, which frames intelligent behavior as the result of feedback-driven control, as well as from cognitive theories that emphasize modularity and coordination. In particular, Global Workspace Theory (GWT) \cite{baars1988,dehaene_naccache_2001,baars_franklin_2003} models cognition as the interaction of multiple specialized processes that compete for access to a limited shared workspace; information that enters this workspace is broadcast and used to coordinate system-wide behavior.

Guided by this perspective, a central design principle of our system is the explicit separation of different types of knowledge and state into distinct, machine-readable representations. We maintain dedicated structures for persistent knowledge and constraints, environmental state, accumulated experience, and short-term memory. Rather than interpreting these components as a biological model, we treat them as a system-level analogy: specialized information stores that are selectively integrated into a shared, task-relevant representation through a controlled coordination step. This design enables transparent state tracking, supports iterative verification against evidence, and allows the system to adapt its actions without conflating long-term knowledge with transient experiences.

Building on these ideas, we introduce Self-Reflective Chain-of-Thought (SRCoT), a framework that models LLM inference as a closed feedback loop over structured internal states. The framework operationalizes reasoning through explicit claim–evidence representations, systematic identification of information gaps and contradictions, and confidence-driven control decisions that determine whether the system should continue generation, revise prior outputs, retrieve additional information, or abstain. Importantly, the separation of memory, experience, and action allows these processes to be coordinated without interfering with one another, making the overall reasoning process more interpretable and robust.

Our main contributions are as follows:

Firstly, we propose SRCoT, a modular reasoning architecture for LLMs that explicitly separates heterogeneous knowledge and state types—environment state, accumulated experience, persistent knowledge, and short-term memory—into distinct, machine-readable representations. This design is functionally inspired by Global Workspace Theory and enables controlled integration of specialized information into a shared, task-relevant reasoning context.

Secondly, we introduce a closed-loop coordination mechanism for chain-of-thought generation in which structured states are selectively combined, evaluated against evidence and constraints, and iteratively updated. This mechanism allows reasoning, memory, and action selection to be decoupled, enabling systematic verification, confidence-based control, and revision without conflating long-term knowledge with transient task experience.

Thirdly, we implement and examine SRCoT in two synthetic settings---emergency medical services and Munich-Sim, an emergency-operations simulation. Selected reports illustrate explicit gap marking, source-labeled outputs, and failure modes under local deployment constraints. These qualitative observations do not constitute a controlled comparison with baseline generation.

\section{Related Work}
\subsection{Self-Reflection, Retrieval, and Metacognitive Control in LLMs}
Recent work has explored mechanisms that allow language models to reflect on, critique, and revise their own outputs in order to improve factuality and robustness. A prominent example is Self-RAG \cite{asai_selfrag_2023}, which frames retrieval, generation, and critique as a learned policy optimized through self-reflection. In Self-RAG, the model decides when to retrieve external documents, generates an answer, and then critiques its own output to guide further retrieval or revision. While effective, this process is realized within a largely unified model context, where reasoning, evidence assessment, and control decisions are implicitly entangled.

In contrast, our approach differs at the architectural level. Rather than learning a unified self-reflection policy, SRCoT explicitly separates heterogeneous knowledge and state types into distinct, machine-readable representations (e.g., persistent knowledge (long-term memory), accumulated experience, environment state, and short-term memory). Control decisions operate over these structured states in a closed feedback loop, enabling verification and revision without conflating transient reasoning outcomes with long-term knowledge. As a result, self-monitoring in our framework is not an emergent behavior of a single model, but a system-level coordination mechanism.

Our work is also related to research on metacognition in neural systems, particularly the idea that intelligent agents should be able to assess their own uncertainty, detect knowledge gaps, and regulate behavior accordingly—often summarized as “knowing what you do not know.”\cite{paul_metacognitive_uncertainty_response_2015, fleming_dolan_neural_metacognition_2012} Prior work operationalizes such capabilities through confidence estimation, self-consistency, verifier models, or reflection prompts. SRCoT can be viewed as a structural instantiation of these ideas: uncertainty and conflict are tracked explicitly in the experience state and used as control signals for action selection (e.g., retrieval, revision, or abstention), rather than being implicitly encoded in model logits or prompt text.

Finally, our approach differs from self-training and self-improvement methods, where models iteratively generate pseudo-labels or revise outputs to update their parameters. SRCoT does not modify model weights and does not treat generated outputs as new training data. Instead, learning is replaced by state accumulation: experiences are stored separately from persistent knowledge and influence future actions through control logic rather than parameter updates. This design choice is particularly important for edge deployments, where continual training is often infeasible or undesirable.

In summary, while SRCoT is conceptually related to self-reflection, metacognition, and self-training, it contributes a distinct perspective by emphasizing explicit state separation and closed-loop coordination as a foundation for reliable and evidence-based reasoning.

\subsection{GWT-inspired state and knowledge representations}
\label{subsec:gwt_json_memories}

A central design motif of our system is the explicit separation of different types of knowledge and state into separate, machine-readable representations (e.g., \texttt{environment.json} (E), \texttt{experiences.json} (X), \texttt{knowledge.json} (K), \texttt{memory.json} (M)). This structure is functionally inspired by Global Workspace Theory (GWT), which describes cognition as the interplay of many specialized processes that compete for access to a limited shared ``workspace''. Contents that enter this workspace can be broadly \emph{broadcast} and thereby coordinate other processes \cite{baars1988,nakanishi2025gwt}. We do not interpret the JSON\footnote{JSON (JavaScript Object Notation) is a lightweight, human-readable data format for structured records using key--value pairs and arrays. In our setting, it serves as a simple container to store many small text items.} files as a biological model, but as a \emph{systemic analogy}: they form specialized knowledge stores that are transferred, in a controlled step, into a shared representation that is relevant for answering.

The split into multiple files serves a practical purpose: it enforces clear semantics about \emph{which type} of information is being used at any given time. For example, \texttt{environment.json} can verbalize situational conditions and constraints (e.g., organizational structure, resource availability, or information about objects and items); \texttt{knowledge.json} can contain relatively stable, curated domain information; \texttt{experiences.json} can collect case examples, decisions, and their outcomes (``what has worked in practice?''); and \texttt{memory.json} can store short-term, task-specific notes or session states. In practice, humans also separate knowledge into categories such as context, general knowledge, and experience/memory, and use different ``access paths'' depending on the task. Through the explicit data structure, this categorization becomes operationalizable for the system: modules can query only those sources that are relevant for a given step, and conflicts (e.g., between stable knowledge and local rules) become visible instead of being implicitly mixed.

Importantly, the JSON structure is not merely ``storage'', but part of the control mechanism. Once contents are transferred into the shared workspace (in our case, for example, a claim--evidence or workspace representation), they can be checked by monitoring and control components. The workflow thus follows a metacognitive pattern: an object level generates and collects content, while a meta level verifies whether constraints are satisfied and whether evidence or context is missing, and then intervenes in a controlling manner \cite{nelsonnarens1994}. The separated stores substantially facilitate this control because checking rules are easier to formulate (e.g., ``local procedures take precedence over general knowledge'', ``environment constraints must not be violated'', ``experience may only count as a heuristic, not as evidence'').

\subsection{Self-reasoning in local, smaller edge LLMs}
\label{subsec:edge_llm_self_reasoning}

A motivation for self-reasoning on local LLMs is that resource limits constrain the context and computation available for evidence checking. An external controller can organize generation into smaller, inspectable steps instead of a single completion. The following mechanisms are design rationales, not measured advantages of smaller models; this study does not compare model sizes.

We consider four mechanisms relevant to such edge settings:
\begin{enumerate}
  \item \textbf{External structure reduces cognitive load.}
  When context, knowledge, experiences, and working memory are separated and structured, the model has to keep less ``in its head'' and mixes less implicitly. This is intended to reduce the mixing of irrelevant information and the omission of local rules. The workspace becomes smaller, clearer, and easier to validate \cite{baars1988}.

  \item \textbf{Iteration turns one-shot generation into error correction.}
  One-shot generation is brittle: an early error propagates through the entire text. In a control loop, errors can be detected and corrected later (e.g., via targeted follow-up retrieval, conflict marking, or conservative phrasing). Whether smaller models benefit disproportionately, and whether their checks are adequate, requires a separate comparative evaluation.

  \item \textbf{Self-monitoring enforces evidence-bound writing.}
  Many hallucinations arise because the model unconsciously ``keeps narrating'' when support is missing. A monitor that explicitly asks ``Which claim is supported by what?'' and marks gaps is intended to discourage this continuation, although the monitor can itself fail. The meta level can then decide to retrieve more, declare uncertainty, or stop \cite{nelsonnarens1994,ji_hallucination_survey_2023}.

  \item \textbf{Local compatibility acts as a hard constraint.}
  In edge setups, the target is often concrete local standards (standard operating procedures, organizational logic, regional medications/protocols in the emergency medicine example). These constraints can be interpreted as the \emph{reference signal} of the control loop: an answer is only acceptable if it matches the local boundary conditions. This constrains the intended output space, but neither model size nor the presence of local rules guarantees a locally valid answer \cite{wiener1948cybernetics,astrommurray2008feedback}.
\end{enumerate}

The proposed role of self-reasoning is therefore \emph{process control}: structured state, iterative checking, and controlled stopping decisions. These mechanisms make aspects of generation inspectable; they do not establish operational deployability.

\subsection{Terminology: Self-Reflection vs.\ Self-Reasoning}
In this paper, we use \emph{self-reasoning} as an umbrella term for multi-step generation that includes intermediate computation, retrieval, and revision. We reserve \emph{self-reflection} for the \emph{meta-level} part of self-reasoning: explicit monitoring of the system's own draft with respect to evidence, constraints, and missing critical information, followed by a control action (e.g., re-retrieval, revision, or abstention). Under this definition, SRCoT's Pass~1 (Draft + Follow-ups) implements \emph{reflection as gap detection} (``what is missing for a justified answer?''), while Pass~2 implements \emph{reflection as controlled revision} (rewrite under the expanded context). The optional Judge step makes this reflective layer externally observable by producing claim--evidence and missing-facts structures. Consequently, ``self-reflective'' in SRCoT does not claim an internal cognitive faculty of the model; it denotes a system-level mechanism that operationalizes metacognitive monitoring and control over an evidence-constrained workspace.

\section{Methodology}
Expanding on the overview presented already in Section \ref{Introduction}, we now detail SRCoT's complete formulation, including its architectural framework and components, as well as its implementation. 

\subsection{Problem Formulation}
We consider evidence-oriented generation under local operational constraints. Let $q$ denote a user query and let $\mathcal{D}$ be a local document collection (e.g., SOPs\footnote{Standard Operating Procedures (SOPs): locally defined, organization-specific operational procedures that translate policies or guidelines into concrete, auditable steps. In this paper, the term includes both medical SOPs/checklists (e.g., assessment, handover, contraindication checks) and non-medical operational rules such as fire brigade Standard Operating/Response Rules (SER; \emph{Standard-Einsatz-Regeln}) used to standardize incident tactics and coordination.}, checklists, organizational rules). The system produces an output $y$ together with an auditable trace $\tau$ (e.g., retrieved snippets, claim--evidence links, detected gaps/conflicts). We model generation as an iterative process over an explicit system state
\[
S_t=\langle E, K, X, M \rangle,
\]
where $E$ encodes environment constraints (local availability, roles, allowed actions), $K$ curated stable knowledge, $X$ accumulated experience heuristics, and $M$ short-term task memory.

A key objective is to reduce unsupported or locally invalid statements. Let $V(y,\mathcal{C},\mathcal{D}) \in \{0,1\}$ be a validator that checks whether $y$ satisfies acceptance criteria $\mathcal{C}$ (e.g., ``claims must be supported by retrieved evidence'', ``local Standard Operating Procedures override general knowledge'', ``do not propose actions unavailable in $E$''). We aim to produce outputs that satisfy $V=1$ while keeping the process resource-bounded on edge devices:
\[
\text{find } y \text{ s.t. } V(y,\mathcal{C},\mathcal{D})=1 \quad \text{using at most } T \text{ iterations and a fixed compute budget}.
\]
When validation cannot be achieved, the system should fail safely by abstaining or asking for missing critical information.
\subsection{Framework}
SRCoT implements a closed-loop inference procedure (Figure~\ref{fig:srcot}). Each iteration alternates between (i) retrieving local evidence, (ii) constructing a structured workspace that separates claims, evidence candidates, and open gaps, and (iii) applying a controller that decides whether to finalize, revise, retrieve more targeted evidence, or abstain. The key design goal is auditability: the system produces not only an answer but also a trace that exposes which constraints were active and which evidence supported each claim.

\vspace{5mm}
\begin{center}
\includegraphics[width=1.0\linewidth]{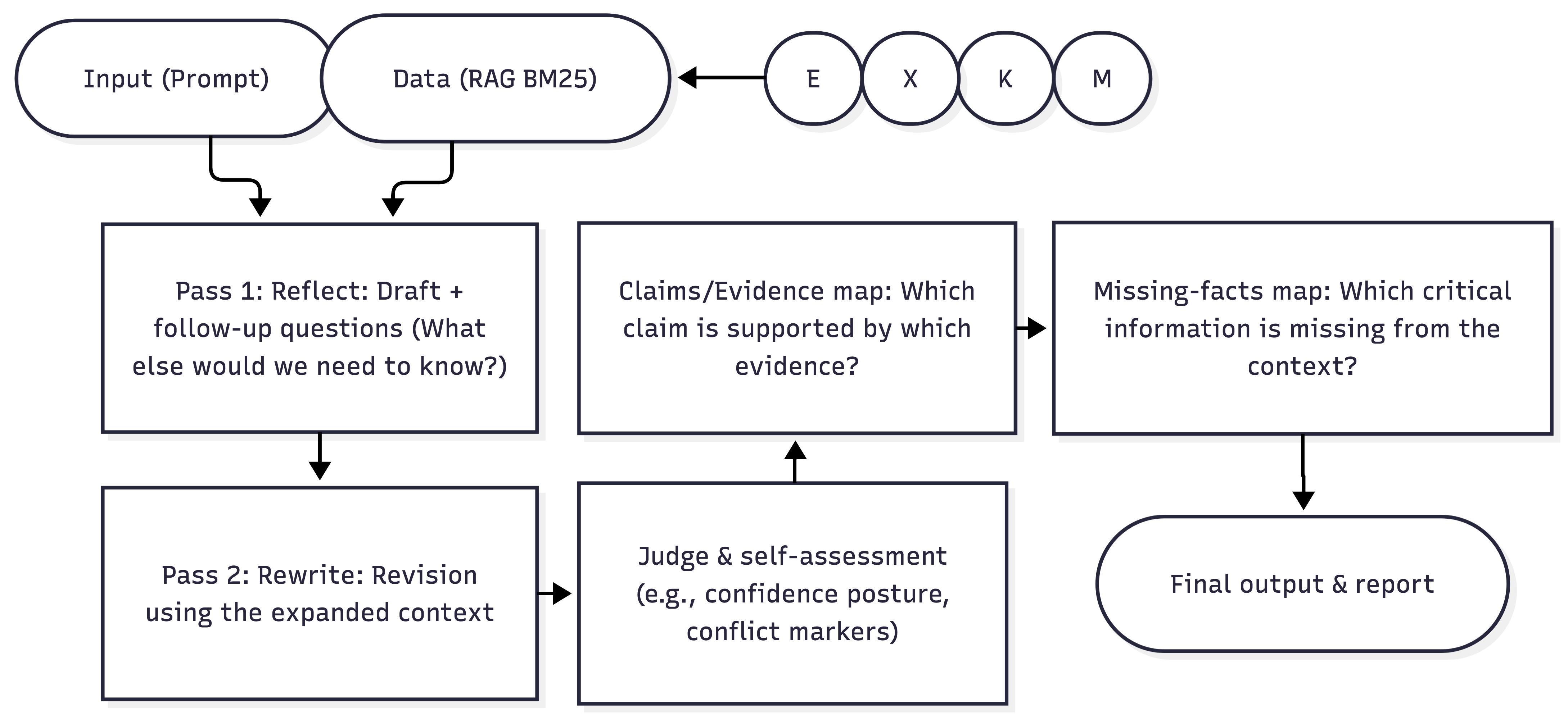}
\captionof{figure}{SRCoT workflow. Source: Author's own illustration.}
\label{fig:srcot}
\end{center}
\vspace{5mm}

\subsection{Workspace}
The framework maintains multiple machine-readable stores---\texttt{environment.json} (E), \texttt{knowledge.json} (K), \texttt{experiences.json} (X), and \texttt{memory.json} (M)---which are selectively integrated into a task-specific workspace $W_t$. The workspace is not merely additional context text; it is a structured representation that contains: (i) atomic claims (statements or recommended steps), (ii) evidence candidates (retrieved snippets labeled with their source identifiers so statements can be traced back to specific documents), (iii) explicit constraints derived from $E$, and (iv) a list of missing critical facts.

This separation serves two practical functions. First, it reduces implicit mixing of heterogeneous information (e.g., preventing heuristics in $X$ from being treated as hard evidence). Second, it enables deterministic checks and prioritization rules (e.g., SOP precedence) to be applied transparently, making failure modes easier to diagnose.
\subsection{Judge and Controller}
After an initial draft is generated, an optional \emph{Judge} step decomposes the draft into checkable claims and attempts to attach local evidence to each claim via targeted retrieval queries. For example, a draft sentence such as \emph{``Treat nausea with dimenhydrinate''} is split into claims (e.g., \emph{indication} and \emph{contraindication checks}); the Judge retrieves the matching local SOP snippet(s) and records a claim--evidence map (e.g., C1 $\rightarrow$ [Knowledge:47 Antiemese SOP]), lists missing critical facts (e.g., age/allergies not specified), and flags conflict markers (e.g., ``contraindication'' / ``do not use'' German: ``Kontraindikation'' or ``nicht anwenden'') that can trigger conservative control actions (retrieve more / ask-human / abstain).

A deterministic \emph{Controller} then selects the next action based on these signals. Typical actions include: \textsc{Finalize} (if coverage and constraints are satisfied), \textsc{Targeted-Retrieve} (if evidence is missing), \textsc{Revise-with-Evidence} (if new snippets were found), or \textsc{Abstain/Ask-Human} (if critical gaps remain). Importantly, the Judge is treated as a tool for producing auditable diagnostics rather than as a ground-truth verifier.
\subsection{Implementation}
We implement SRCoT as a lightweight Python pipeline designed for local (edge) OpenAI-compatible inference servers (e.g., LM Studio\footnote{LM Studio is a local application that hosts language models on a personal computer and exposes an API compatible with OpenAI-style chat interfaces, enabling offline experiments without cloud calls.} with OpenAI GPT-oss-120b\footnote{Model names often include an approximate parameter count; ``120b'' indicates a model with on the order of 120 billion parameters, which generally correlates with higher compute and memory requirements. GPT-oss-120b reasoning content originates from the model’s output format (chain-of-thought tokens).}). Retrieval is performed with BM25\footnote{Okapi BM25 is a lexicon-based probabilistic ranking function widely used in information retrieval. It scores documents by term overlap with length normalization and term-frequency saturation that provides a strong, deterministic baseline for sparse text retrieval.}  \cite{robertson_zaragoza_bm25_2009} over pre-segmented JSON corpora\footnote{Here, ``JSON corpora'' refers to locally stored, pre-segmented text items represented as JSON objects (e.g., one procedure, rule, log entry, or note per item) with optional metadata fields. BM25 is applied to the textual content fields, while metadata is used for provenance labels and reporting.}. The system supports a two-pass workflow (draft and revise) and optionally adds a Judge step plus a targeted retrieval--revision loop. Each run produces a structured report containing retrieved contexts, intermediate drafts, follow-up questions, and (if enabled) claim--evidence maps and control decisions. This design is intended to be modular, inspectable, and feasible under edge constraints (limited context window and compute).

The state, validator, and controller descriptions above specify the conceptual control-loop view; the implemented prompt contracts and optional Judge remain fallible approximations of these design objectives.

\section{Experimentation}
\label{sec:experimentation}

We evaluate SRCoT as an \emph{exploratory prototype} using synthetic, trace-rich case studies. The goal is not to establish superiority over alternative methods, but to document (i) feasibility on edge hardware, (ii) auditability and trace quality, and (iii) typical failure modes and control-loop behaviors that motivate future systematic experiments.

All runs follow the SRCoT reporting convention: the system produces a final answer plus a structured trace consisting of retrieved context blocks (labeled by store provenance), an intermediate draft (Pass~1 ``Reflect''), and a revised output (Pass~2 ``Finalize''). Whenever information is missing from the provided context, the output must explicitly mark it as \texttt{not specified}. For some tasks, SRCoT additionally generates a focused second retrieval query (\texttt{query2}) derived from follow-up questions or gap signals, enabling a retrieval--revise loop. Optionally, a \emph{Judge} step produces a claim--evidence map and enumerates missing critical facts, conflicts, and a conservative confidence estimate.

We treat the SRCoT report as the primary evaluation artifact. Accordingly, we emphasize \emph{trace-based proxy criteria} that can be inspected without medical ground truth labels: (i) evidence anchoring (presence of source keys for key statements), (ii) constraint adherence (e.g., ``no invented values''), (iii) explicit gap marking (\texttt{not specified} instead of completion), and (iv) controller recommendations (retrieve more / revise / abstain / dispatch suggestion).

\subsection{Case Study 1: EMS -- local SOP as a reference signal in the SRCoT control loop}
\label{subsec:case_ems}

This case study illustrates SRCoT in a setting where local protocols dominate, resources and responsibilities vary by situation, and unsupported specifics are hazardous. The presentation is intentionally \emph{abstract} (no dosages or concrete therapy instructions) to demonstrate the system mechanism rather than provide medical guidance.

An edge deployment in emergency medical services (EMS) should support responders with documentation/checklists and with a structured derivation of next steps under local constraints. Local SOPs and availability (e.g., medications, monitoring, transport options) differ across organizations and regions. Our prototype was executed locally on an Apple M4-family MacBook Pro with 128\,GB memory, using a local OpenAI-compatible inference endpoint. The saved EMS report and result metadata identify \texttt{gpt-oss-120b}, whereas the associated request file specifies \texttt{gpt-oss-20b}. We therefore do not use that request file to reconstruct the selected run's settings. The exact model artifact, quantization, and inference-runtime version are not established by the examined case files. This shows that, in principle, the system can run on a lightweight, self-contained device that could be carried in an ambulance or other mobile assets. This is particularly relevant for prehospital settings, because local execution (without cloud dependency) enables offline capability, low infrastructure requirements, and better control over data handling.

SRCoT separates heterogeneous information into four provenance-labeled stores. The \texttt{environment.json} store ($E$) captures operational context such as vehicle type, available devices/medications, responder roles (e.g., EMT/paramedic/physician), transport times, and local escalation pathways. The \texttt{knowledge.json} store ($K$) contains general medical background concepts (e.g., ABCDE/FAST as principles) but explicitly \emph{without} local binding authority. The \texttt{experiences.json} store ($X$) collects practice-oriented heuristics such as recurring documentation pitfalls or frequent contraindication checks (i.e., ``what tends to work in this organization''). Finally, the \texttt{memory.json} store ($M$) represents the current case state, including patient data, vital-sign trends, actions already performed, open questions, and timestamps.

The control loop proceeds in four steps. First, the system \emph{measures} via retrieval: BM25 searches for relevant SOP passages based on the incident description and findings stored in $M$ (e.g., ``stroke algorithm'', ``neurological emergency / FAST checklist'', ``documentation requirements''). Second, it performs a workspace build: \texttt{srcot.py} constructs a workspace $W_t$ that aggregates proposed next steps as claims, attaches referenced SOP passages as candidate evidence, and injects explicit constraints from $E$ (e.g., ``not available'', ``physician-only'', ``transport priority''). Third, the monitor \emph{compares} by applying consistency checks, including SOP priority (local SOP overrides general knowledge on conflict), resource consistency (no actions requiring items marked unavailable in $E$), and evidence binding (claims without SOP/document support are marked uncertain or removed). Fourth, the controller \emph{corrects}: if gaps remain (e.g., SOP not found, incomplete patient data), it triggers targeted re-retrieval (e.g., query rewriting from diagnoses) or downgrades the output policy from ``recommendation'' to ``checklist/questions''.

In EMS, ``globally plausible'' is often ``locally wrong'' (different medications, different escalation paths, different documentation duties). The architecture therefore enforces a hard constraint: \emph{local compatibility} as a reference signal. Experiences ($X$) may \emph{guide} the process (e.g., which checks commonly go missing) but may not replace SOPs as evidence. This is intended to discourage overconfident specifics by replacing unsupported elaboration with explicit gap marking; the case study does not measure the rate of such failures. If, after $T$ iterations, no evidence-bound, SOP-compliant answer can be produced, the system is designed to output an honest fallback
(e.g., ``SOP source not found / data missing; please consult the local algorithm''). In medical contexts, this controlled abort is an intended safeguard, not a demonstrated guarantee of safety.

\subsubsection{Interpretation}

This EMS scenario illustrates that follow-up questions in a self-reflective setting are not merely a by-product of text generation, but can be interpreted as a \emph{systematic marker} of uncertainty management and metacognitive control. Functionally, this mirrors a common prehospital pattern: an initial action structure is followed by an internal check of which information is actually known and which parameters are missing in order to justify the next steps safely, protocol-compliantly, and in a handover-ready form.
\\\\
In the SRCoT report, the system first produces an SOP-proximal, fully structured checklist for prehospital stroke handling. In parallel, it explicitly marks missing fields (e.g., as \texttt{not specified}) and then generates targeted follow-up questions. Notably, these follow-ups are content-specific: they address precisely those critical parameters that matter for prehospital decision-making and handover logic and were visibly missing in the preceding output—namely heart rate and temperature (standard monitoring values that remained unclear), blood glucose (required to rule out hypoglycemia, but not given), intravenous (IV) access status (required but not stated unambiguously), and symptom onset time (a time-critical key variable for stroke pathways that was missing).

This makes a metacognitive sequence reconstructable from a single report: the system can respond in a structured, action-guiding manner while simultaneously recognizing that core information is missing for safe application and credible handover, and it then ``translates’’ these gaps into \emph{concrete, operational follow-up questions} (information seeking instead of free completion).

For the argument of this paper, this suffices as a concrete case: follow-up questions form the \emph{operational interface} between knowledge application (object level) and meta-control (meta level) \cite{nelsonnarens1994}. Functionally, this corresponds to confidence-controlled information seeking: as long as critical variables are missing, the system does not ``keep elaborating'' but switches its generation policy toward data collection/clarification \cite{kepecs2008,flemingdaw2017,pouget2016}.

The report provides qualitative evidence that follow-ups in this setting are \emph{not arbitrary} but \emph{gap-driven}: visible missing facts are converted into specific questions (missing facts $\rightarrow$ targeted questions). This makes follow-ups an observable, auditable indicator of self-monitoring: the system exposes \emph{where} its evidence base ends and what additional information would be required to make the next decision robust.

Beyond this, the same artifact trail (output + marked gaps + follow-ups + retrieval blocks + Judge map, if enabled) supports further \emph{qualitative} claims. The key is to frame these as properties of the implemented control procedure (``becomes visible'', ``is documentable'', ``is auditable'') rather than as strong claims about clinical correctness or model ``cognition''.

\subsubsection{Key findings}

\begin{itemize}
  \item \textbf{Evidence discipline as an explicit writing contract (auditability over plausibility).}
  By combining the \texttt{not specified} convention with a minimum citation requirement, a central LLM failure mode is \emph{systemically discouraged}: plausible embellishment in the absence of data. This is qualitatively supported because (i) missing values remain visible and (ii) central steps/claims are tied to concrete retrieval blocks (inline citation keys). This does not make the output ``true'', but it makes it \emph{inspectable}: readers can verify which local text anchor served as the basis for each action element.

  \item \textbf{The object-level vs.\ meta-level separation is visible in the artifact.}
  In the two-phase process (draft/follow-ups $\rightarrow$ final), the report shows a role separation: Pass~1 produces a provisional action structure while diagnosing incompleteness; Pass~2 integrates additional evidence and stabilizes the formulation. This sequence is compatible with the Nelson \& Narens distinction (object level: task performance; meta level: monitoring/control) without requiring any claims about ``consciousness'' \cite{nelsonnarens1994}.

  \item \textbf{Local compatibility becomes visible as a hard constraint (constraints $\rightarrow$ language form).}
  When \texttt{environment.json} and local SOP fragments appear in context, it becomes qualitatively visible that the response is not primarily oriented toward ``global'' medical text, but toward locally coded terminology, role logic, and handover structure. This is not a guarantee, but it is an auditable trace: the used terms and steps can be found in the cited local blocks rather than remaining purely generic.

  \item \textbf{The Judge operationalizes ``self-critique'': claim decomposition and evidence coverage.}
  If the Judge is enabled, it becomes possible to show qualitatively that self-monitoring appears not only as follow-up questions but as a \emph{structured verification} of the final answer: the output is decomposed into claims; targeted queries are generated per claim; evidence coverage is recorded as a map (\texttt{claim\_evidence\_map}). As a result, ``confidence'' is not emitted as an intuition but is tied to observable proxy signals (e.g., coverage, missing critical facts, conflict markers).

  \item \textbf{A closed control loop is describable as a system procedure (cybernetics as an objective frame).}
  Without making neurobiological claims, one can show qualitatively that the procedure matches the minimal structure of a feedback controller:
  (i) measurement/observation via retrieval,
  (ii) comparison/diagnosis via validator/Judge signals (gaps, conflicts, missing evidence),
  (iii) corrective action via policy (e.g., \texttt{Targeted-Retrieve}, \texttt{Refocus}, \texttt{Abstain}),
  (iv) re-measurement and revision.
  The key point is the \emph{documented causal trace}: diagnostic artifacts (gap/conflict/coverage) are visible in the report and can be linked to the triggered control action. This makes the cybernetic interpretation procedural rather than metaphorical \cite{wiener1948cybernetics,astrommurray2008feedback,ashby1956}.

  \item \textbf{A functional analogy to conflict monitoring is admissible if stated as a proxy.}
  Even without claiming that an ACC mechanism is ``reproduced'', one can show qualitatively that the system uses conflict/risk hints as \emph{triggers for control adaptation}: the appearance of conflict markers or contraindication hits $\rightarrow$ a more conservative confidence posture $\rightarrow$ additional information seeking or abstention. This is compatible with conflict-monitoring accounts (conflict $\rightarrow$ increased control) but should be framed explicitly as an \emph{algorithmic analogy} \cite{newmanmcgaughy2015}.
\end{itemize}

The scenario documents a gap-driven control logic: workflow-critical parameters that are missing in the trace are explicitly marked as missing and then translated into concrete follow-up questions rather than replaced with plausible completions (e.g., ``What is the current heart rate?'', ``What temperature was measured?'', ``What is the bedside glucose value?'', ``Is IV access established?'', ``When did symptoms begin?''). In combination with citation obligations and an explicit missing-value convention, this yields an auditable output that exposes local evidence anchors and enables post-hoc verification of individual steps. When the optional Judge is enabled, self-monitoring becomes a second verification layer (claim decomposition, evidence coverage, gap/conflict signals) that can deterministically trigger more conservative control actions (targeted retrieval/abstention). Overall, the EMS case is informative because the causal trace remains explicit in the report: missing critical parameters are marked and immediately transformed into follow-up questions (``not specified'' $\rightarrow$ query) rather than being hidden in filler text.

\subsection{Case Study 2: Munich Sim as a synthetic testbed for SRCoT}
\label{subsec:usecase_munich_sim}

To stress-test SRCoT beyond small hand-curated corpora, we use \emph{Munich Sim (München Einsatz-Simulation)}, a software-based emergency operations simulation (built with GPT 5.2) that generates large volumes of structured and semi-structured operational text. Munich Sim provides a controllable environment in which SRCoT can be evaluated under semi-realistic documentation and dispatch dynamics while avoiding sensitive real-world data. Concretely, the simulator produces: (i) incident overviews (e.g., location, incident type, assigned units, and a mechanically derived dispatch proposal), (ii) log-style event streams (status changes, control room updates), (iii) radio-like message snippets, and (iv) continuously updated resource and material inventories. These artifacts are ingested into SRCoT's separated stores (e.g., \texttt{environment.json} for rules/resources/AAO\footnote{AAO-like refers to an AAO (German: Alarm- und Ausrückeordnung), i.e., the predefined dispatch/response plan that specifies which units are sent for a given incident type (here used as a simplified, rule-based dispatch proxy in the simulator).} logic, \texttt{memory.json} for incident-specific timelines and messages), enabling retrieval and verification against a growing, heterogeneous text base. A key objective is feasibility on local edge deployments. In our prototype runs, we executed a large local model (\texttt{gpt-oss-120b}) on an Apple laptop with 128\,GB RAM. Even with per-request compute times on the order of minutes, SRCoT produced long, structured, and source-anchored reports. While latency is a limitation for real-time use, these runs demonstrate that high-traceability reporting is possible without cloud dependency, which is relevant for offline-capable or privacy-preserving deployments.
\vspace{5mm}
\begin{center}
\includegraphics[width=0.9\linewidth]{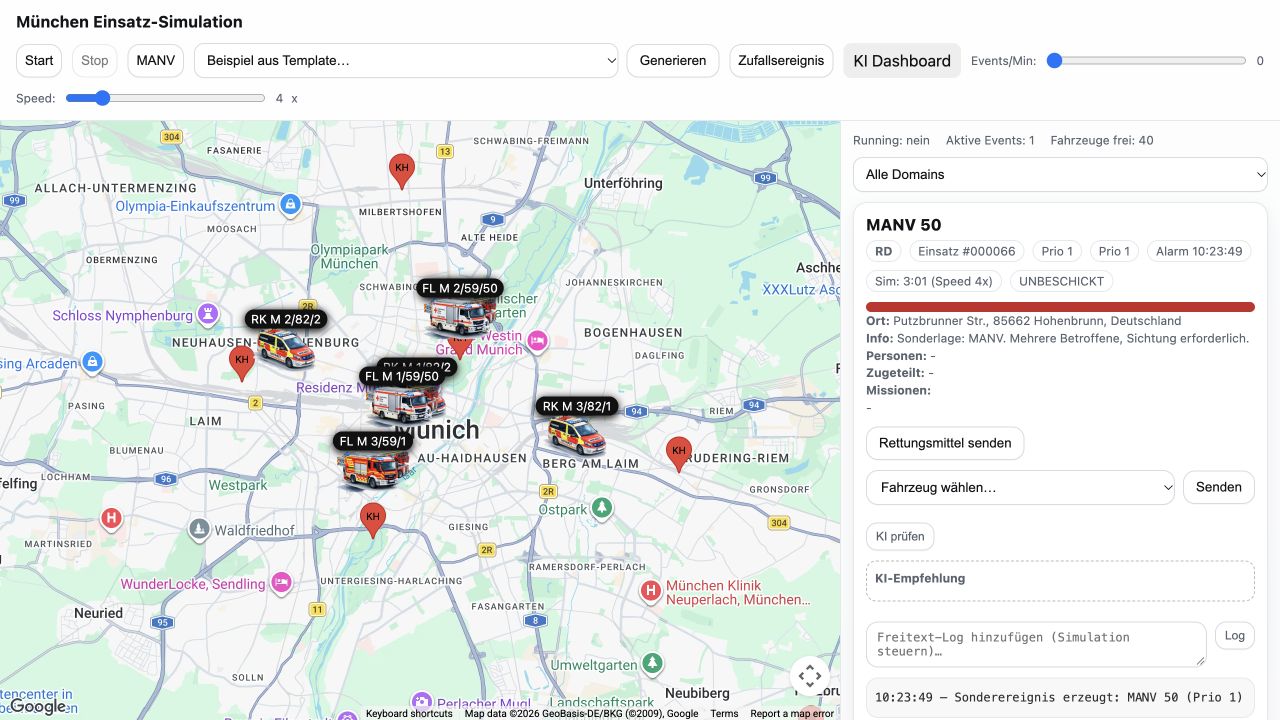}
\captionof{figure}{Munich Sim: illustrative user-interface view captured on 14 October 2026, not a recorded evaluation trace. Simulation software by the authors; vehicle icons generated using an OpenAI GPT model. Map: Google Maps. Map data \copyright{}2026 GeoBasis-DE/BKG (\copyright{}2009), Google.}
\label{fig:munichsim}
\end{center}
\vspace{5mm}

\subsubsection{Tasks}
\label{subsec:munich_tasks_examples}

We focus on three task families that directly exercise SRCoT's closed-loop design. For each, we include a compact excerpt from a representative SRCoT report and summarize what it reveals about the system's behavior.

\paragraph{(T1) Inventory and material accounting from contextual tables}
\label{subsubsec:munich_t1_material}

SRCoT is prompted to compute cumulative material consumption from a context-provided inventory overview and to rank top consumers. The prompt enforces the rule: if units or quantities are incomplete, output must be \texttt{not specified}. This task probes whether SRCoT can extract quantities reliably, avoid hallucinating missing values, and present results in an auditable format.

In the exemplar report, SRCoT outputs a tabular summary of the total material consumption and explicitly preserves the item identifiers and units (e.g., \texttt{ml}, \texttt{pcs}, \texttt{pairs}). Top consumers are ranked by absolute amount. Crucially, one item is not fully specified in the context and SRCoT marks it accordingly:
\begin{lstlisting}[basicstyle=\ttfamily\small,frame=single]
con_fuel_mix -- 2200 ml;
con_hand_disinfectant -- 1250 ml;
con_saw_chain_oil -- 300 ml;
...
con_absorbent -- 10 ... not specified (truncated as "10 ...") [Environment:8].
\end{lstlisting}

This exemplar illustrates a concrete effect of SRCoT's ``missing-value contract'': instead of completing truncated quantities with plausible numbers, the model is steered toward explicit uncertainty marking. For operational contexts (logistics, resupply, cost accounting), this is valuable because it produces \emph{audit-friendly} outputs: all numeric values can be traced to a single inventory source block, while incomplete entries remain visibly incomplete.

\paragraph{(T2) Under-coverage detection and dispatch-style control output (AAO proxy)}
\label{subsubsec:munich_t2_coverage}

SRCoT is prompted to check whether an incident is under-dispatched relative to the situation development (logs/radio) and an AAO-like mechanical proposal. The output format is constrained: the ``Direct Answer'' must start with a one-line JSON object indicating recommended unit types, a short reason, and a confidence value. Duplicates in the unit array represent multiplicity.

In the exemplar report, SRCoT produces a JSON control recommendation and justifies it by citing both (i) the incident overview (unassigned incident) and (ii) the AAO proposal\footnote{German EMS unit abbreviations: RTW ("Rettungswagen") is an ambulance staffed for patient transport and prehospital care; NEF ("Notarzteinsatzfahrzeug") is a rapid-response physician unit that brings an emergency doctor to the scene but typically does not transport the patient.}:

\begin{lstlisting}[basicstyle=\ttfamily\small,frame=single]
{ "unitTypes": ["RTW","NEF","RTW"],
  "reason": "Under-coverage: no units assigned while AAO proposes RTW and NEF" ...}
  Incident is unassigned (Assigned: none) [Environment:7];  
  AAO proposal lists RTW, NEF, RTW [Environment:7]. 
  No logs/radio confirm already assigned units [Environment:9-10].
\end{lstlisting}

This exemplar demonstrates SRCoT's control framing in a dispatch-like setting: the system detects an explicit state mismatch (``proposal exists'' vs.\ ``no assignment'') and emits a structured control output (JSON) suitable for downstream tooling.

\paragraph{(T3) Incident documentation synthesis from event streams (logs and radio)}
\label{subsubsec:munich_t3_documentation}

SRCoT is prompted to generate a detailed and factual incident documentation (narrative + bullet points) using \emph{only} logs/radio/context. Constraints include: (i) no medical dosage or therapy instructions, (ii) no invented timestamps, and (iii) missing items must be marked as \texttt{not specified}.

In a representative Munich Sim incident (``apartment fire with missing person''), SRCoT generates a structured documentation with explicit sourcing of core statements, including: location/incident overview, AAO-like proposal, events such as ``fire out'' and ``ladder unit released'', an explicit mention of missing arrival time, and transport/hand-over events \footnote{German fire service unit types: HLF (\emph{"Hilfeleistungslöschfahrzeug"}) is a multi-purpose fire engine with technical rescue equipment; DLK (\emph{"Drehleiter (automatisch) mit Korb"}) is an aerial ladder platform for access/rescue and firefighting; ELW (\emph{"Einsatzleitwagen"}) is a command vehicle used for incident coordination and communications.} \footnote{Glasgow Coma Scale (GCS): a standardized neurological assessment of consciousness based on eye opening, verbal response, and motor response (total score 3--15)}:

\begin{lstlisting}[basicstyle=\ttfamily\small,frame=single]
Location and incident overview [Environment:7].
AAO proposal: HLF, DLK, ELW [Environment:7].
"Feuer ist aus, Brandrauch entfernt. DLK entlassen ..." 
[Memory:253 admin_injected; Memory:503 #000053 Memory].
Arrival time not present -> not specified.
Transport and handover documented (e.g. GCS and temperature) 
[Memory:231 Vor Ort; Memory:235 Vor Ort].
\end{lstlisting}

\subsubsection{Stress-Test Scenarios (Munich Sim)}
\label{subsec:munich_case_studies}

Beyond the core tasks above, we include three stress-test scenarios that explicitly manipulate the simulated state. They are used to probe SRCoT's closed-loop behavior under disturbance and scale, rather than to claim factual correctness.

\textbf{Stress-Test Scenario A (Munich Sim): Large-memory incident report generation under injected control inputs}

This case is retained as a \emph{stress test} rather than a clean accuracy benchmark. In early prototype runs, the incident workspace (especially \texttt{memory.json}) becomes very large, and we observed occasional provenance mistakes in the narrative layer (i.e., statements that originate in \texttt{Memory:*} may be attributed inconsistently in some report passages). We therefore do \emph{not} interpret this scenario as evidence of superior factual correctness. Instead, it documents two properties central for SRCoT in practice: (i) the ability to build structured documentation from a massive, heterogeneous event stream, and (ii) controllability of scenario escalation via external interventions.

We actively manipulated and escalated the scenario by repeatedly injecting control-room updates (``Leitstelle'' messages) into the incident timeline. These injections are an explicit experimental intervention: they modify the observed system state and force SRCoT to re-contextualize dispatch and documentation decisions. Concretely, the simulation is steered by messages such as escalation to a mass-casualty setting (German "MANV") from an initial house fire incident \,[Memory:25 admin\_injected; Memory:68 admin\_injected; Memory:70 admin\_injected], confirmation of an "Atemschutznotfall" (breathing apparatus emergency) \,[Memory:66 admin\_injected], and later updates about triage results and transport needs \,[Memory:142 admin\_injected; Memory:265 admin\_injected]. This tests whether SRCoT remains usable when the state evolves through abrupt, externally injected updates rather than smooth progression.

Even under this high-noise setting, SRCoT produces a structured incident report (narrative + bullet points) that integrates core developments and marks missing items explicitly. The key point is not the medical content but the \emph{traceable reconstruction of events} from \texttt{memory.json}:

\begin{lstlisting}[basicstyle=\ttfamily\small,frame=single]
Situation updates (dispatch-relevant):
- Wohnung in Vollbrand, Hausflur verraucht ... Menschenrettung eingeleitet
  (apartment fully involved, corridor smoke-logged 
  ... rescue of people initiated)
  [Memory:37 admin_injected]
- Feuer ist aus, Brandrauch entfernt. DLK entlassen ...
  (fire is out, smoke removed/ventilated. Aerial ladder released ...)
  [Memory:253 admin_injected; Memory:503 #000053 Memory]
- RTW transportiert ... Atemschutznotfall ... keine weiteren RTWs benötigt
  (ambulance transporting ... BA/SCBA emergency 
  ... no further ambulances required)
  [Memory:234 admin_injected; Memory:484 #000053 Memory]
Escalation signals:
- MANV 50 (initial triage estimate) [Memory:25; Memory:68; Memory:70]
- revised counts after re-sighting [Memory:84; Memory:265]
Field observations:
- "Vermisster weiterhin nicht gefunden." (missing person still not found)
  [Memory:92--118 Vor Ort]
Transport / handover:
- transport preparation and hospital handover noted [Memory:232; Memory:235]
\end{lstlisting}

\subsubsection{Interpretation}

The scenario suggests three closely related takeaways: workspace scale is a first-class evaluation dimension, because even when \texttt{memory.json} contains dozens to hundreds of atomic updates (dispatch injections, radio-style messages, unit status lines), SRCoT can still synthesize a coherent structure without requiring a human to manually merge the full log; injected control-room messages operationalize ``disturbances'' for cybernetic evaluation, since Leitstelle injections perturb the observed state and can trigger downstream control actions (dispatch recommendations, escalation/de-escalation), enabling repeatable stress tests by programmatically steering incident trajectories; and a critical failure mode appears under high load, namely provenance/citation slippage in the narrative layer, which indicates that evidence \emph{binding} (mapping claims to concrete \texttt{Memory:*} keys) is a separate problem from retrieval itself and motivates stricter claim--evidence enforcement (e.g., requiring at least one store key per operational paragraph). Overall, this case supports a conservative claim: even when model reasoning is insufficient for robust correctness guarantees, SRCoT can still generate \emph{structured, inspectable artifacts} from very large operational traces, and the simulation can be systematically manipulated through injected state updates to test controller behavior.

\textbf{Stress-Test Scenario B (Munich Sim): Symptom-triggered SOP retrieval via injected state intervention (``nausea'')}

To test whether SRCoT reliably reacts to \emph{protocol-relevant symptom changes} (rather than producing generic text), we intervened in the incident stream by injecting a minimal symptom statement: the dispatcher log reports that the patient feels nauseous \,[Memory:70]. This intervention is deliberately small (one symptom) but operationally meaningful, because it should trigger a specific antiemesis protocol entry and a contraindication check.

After injecting the symptom, we issued a focused follow-up question (\emph{``was können wir gegen die Übelkeit unternehmen?''}). SRCoT expanded it into a safety-relevant \texttt{query2} that explicitly requests missing critical facts (age, allergies/contraindications) before any protocol-level action is rendered:
\begin{lstlisting}[basicstyle=\ttfamily\small,frame=single]
Query2: "... Gibt es bekannte Allergien oder Kontraindikationen 
gegen Dimenhydrinat?
Wie alt ist der Patient?"
\end{lstlisting}

SRCoT produced a structured checklist that anchors the symptom trigger to [Memory:70] and retrieves the antiemesis protocol (including a contraindication list and a protocol-specific dosage string) from the local knowledge base \,[Knowledge:M: Dimenhydrinat Antiemese -- Dimenhydrinat (S.\ 47)]. Crucially, the run also surfaces that key patient attributes are missing (e.g., age, allergy status), and it propagates these gaps as explicit follow-up questions.

With the optional Judge enabled, the trace exposes a claim-level decomposition and a \emph{missing critical facts} map. In this scenario, the Judge flags contraindications as critical claims (e.g., age $<1$ year, epilepsy, interacting medications, allergy) and enumerates missing variables (age/history) as missing critical facts (M1/M2). This yields an auditable control signal: the system should only proceed if these missing facts are resolved, or it should downgrade to \texttt{not specified} / ask-human behavior.

\subsubsection{Interpretation}

A minimal symptom intervention is already sufficient to test protocol-triggered retrieval: a single injected nausea line \,[Memory:70] pivots the retriever toward the corresponding SOP fragment \,[Knowledge:M: Dimenhydrinat Antiemese -- Dimenhydrinat (S.\ 47)], enabling a controlled stimulus--response chain (state intervention $\rightarrow$ retrieval $\rightarrow$ structured output). In the same scenario, follow-up queries make ``knowing what is missing'' operational: the controller's \texttt{query2} and the Judge's missing-facts map provide a concrete metacognitive control mechanism that requests missing variables instead of completing them. At the same time, the run exposes a safety-critical failure mode that must be blocked by micro-policy: in early prompts, the absence of contraindication information can be implicitly treated as ``assume none''; for safety-critical domains this must be explicitly forbidden, so missing contraindication fields remain \texttt{not specified} and trigger follow-ups or abstention/human escalation. Overall this scenario is a controlled software test of SRCoT's closed-loop behavior: a state change is injected, retrieval should pivot to the correct protocol fragment, and the Judge should expose missing critical facts as control-relevant signals. It is \emph{not} evidence of clinical correctness beyond the local SOP text and the implemented control policies.

\textbf{Stress-Test Scenario C (Munich Sim): Follow-up question that exceeds the simulation AAO (supporting units)}

In our CPR documentation scenario, SRCoT produces a structured incident narrative from logs and dispatch context. During Pass~1 (\emph{Reflect}), it additionally generates follow-up questions that are \emph{not answered} by the available logs, but would be operationally relevant in real EMS workflows. One example is:
\begin{lstlisting}[basicstyle=\ttfamily\small,frame=single]
Follow-up (Pass 1 / Reflect):
  "Gab es zusätzliche Unterstützungseinheiten (z.B. FW-Einheiten),
   die während des Einsatzes angefordert wurden?
   (Were additional supporting units (e.g., fire brigade units)
   requested during the incident?)"
\end{lstlisting}

In this run, the simulated dispatch logic (AAO proxy) initially assigns only \texttt{RTW/NEF}. In real-world resuscitation responses, additional assets are often considered (e.g., transport redundancy, specialized equipment, or parallel arrival optimization via first-responder resources such as \texttt{HLF/DLK/KTW}). In the simulator, requesting such units may not change the world state; nonetheless, this follow-up highlights that SRCoT can surface \emph{operational context gaps} that matter for a complete narrative and handover-quality documentation, even if the simulator does not model the associated benefit.

Crucially, SRCoT does \emph{not} invent unlogged arrivals or requests. Instead, the question is exposed as an explicit uncertainty/gap signal in the trace (Pass~1 follow-ups), which can be used to trigger a targeted re-retrieval step if such information exists elsewhere in the corpus (e.g., radio snippets or later control-room updates).

\subsubsection{Interpretation} SRCoT's gap-driven information seeking can extend beyond ``what is in the logs'': it may generate follow-up questions that reflect operational completeness criteria (``Is there missing context needed for a full report?'') rather than merely paraphrasing retrieved statements. Importantly, the same no-invention discipline can apply to operational context, not only to medical specifics: if the trace does not confirm whether supporting units were requested, SRCoT can surface this uncertainty explicitly instead of silently assuming a typical real-world escalation pattern. In this sense, the follow-up can itself function as a control signal—either as a retrieval/verification trigger (``search for supporting-unit evidence'') or as a documentation-quality checklist item depending on the downstream use case.
Overall, this scenario supports a conservative claim: even when the simulator's AAO proxy is minimal and the logs do not encode certain operational details, SRCoT can still expose \emph{structured, inspectable artifacts} that make missing operational context visible. In this sense, the follow-up question acts as a traceable meta-control signal (``missing context'') rather than as a dispatch recommendation, and it provides a practical handle for either targeted re-retrieval (if such evidence exists) or explicit gap marking in handover-quality documentation.

\subsubsection{Key Findings}
\label{subsec:munich_insights}

Across tasks and scenarios, Munich Sim supports the paper's framing of closed-loop self-reasoning as a \emph{controllable} and \emph{auditable} process: the SRCoT \emph{report} becomes the central evaluation object (audit-first), because it exposes not only the final text but also the provenance trace (store keys) that makes key statements inspectable even without gold-standard labels. In addition, the \texttt{query2} mechanism operationalizes information seeking as a proxy for metacognitive control: detected gaps trigger refined retrieval and revision, transforming one-shot generation into a deliberate ``retrieve--revise'' loop. The simulator also enables iterative governance via micro-policies (e.g., strict JSON schemas, \texttt{not specified} contracts, and content restrictions such as ``no dosage instructions''), which can be tuned on synthetic cases without retraining the underlying model. Finally, the synthetic scale of heterogeneous logs and radio-style streams exposes failure modes early: increased volume and variety amplify drift, omission, and provenance errors, and the stress-test scenario in particular highlights evidence-binding under large \texttt{memory.json} as a distinct bottleneck that motivates stricter claim--evidence enforcement and retrieval prioritization.

\section{Conclusion}
In this work, we introduced \emph{Self-Reflective Chain-of-Thought (SRCoT)}, a modular framework for evidence-based text generation that treats LLM inference as a closed \emph{Retrieve--Reflect--Revise} control loop rather than a one-shot completion. SRCoT combines local retrieval, explicit state separation (environment, knowledge, experiences, memory), and a reflection layer that operationalizes self-monitoring as \emph{auditable artifacts}: claim--evidence links, missing-facts signals, and conservative control actions (retrieve more, revise, ask for clarification, or abstain). The central contribution is not a new model, but a process architecture that makes reliability a property of the \emph{system procedure} and its trace, not a hope placed in fluency.

Because LLMs are trained on human-produced knowledge artifacts, their outputs can mirror patterns of human reasoning while still lacking an explicit mechanism for evidence-bound control. Human cognition relies not only on generating candidate thoughts, but also on metacognitive regulation (monitoring uncertainty, detecting conflicts, and actively seeking missing evidence) \cite{nelson_narens_metamemory_1990, flavell_metacognition_1979}. SRCoT instantiates an analogous \emph{system-level} control layer: it decomposes outputs into checkable claims, binds them to retrieved local evidence, flags contradictions and information gaps, and then selects conservative control actions (targeted retrieval, revision, ask-human, or abstention). In this sense, SRCoT operationalizes self-reflection as explicit monitoring-and-control over an evidence-constrained workspace rather than as an assumed internal faculty.

Across the two case-study families, selected SRCoT traces illustrate the intended shift from plausible completion toward \emph{evidence-bound writing}. In the EMS scenario, the system produced structured, SOP-proximal documentation while explicitly flagging missing workflow-critical variables as \texttt{not specified} and translating them into targeted follow-up questions. In Munich Sim, SRCoT scaled to heterogeneous log and radio streams and produced source-anchored reports, numeric summaries, and dispatch-style control objects; importantly, it also exposed gaps and mismatches (e.g., under-coverage signals) without fabricating absent state. Taken together, these traces show that a closed-loop workflow can expose missing facts and proposed evidence links in local deployments. They do not establish a reduction in unsupported statements or improved constraint adherence relative to a baseline.

A key implication is that SRCoT provides a practical path toward \emph{auditable decision support} under edge constraints. Offline execution reduces cloud dependency and improves control over data handling, while the report-first design enables post-hoc inspection, debugging, and governance: reviewers can see which statements were supported by which local fragments, which critical facts were missing, and which control actions were triggered. In operational contexts where “globally plausible” can be “locally wrong,” the ability to encode local SOPs and resource constraints as a reference signal is central: SRCoT is designed to prefer local compatibility and controlled abstention over generic completion.

At the same time, our results do not imply clinical correctness, outcome improvement, or general superiority over baselines. The present evaluation is exploratory and primarily demonstrates feasibility, controllability, and trace quality. SRCoT turns reliability questions into testable system properties (evidence coverage, binding, gap detection, and control policies) and makes failures visible as concrete artifacts rather than hidden model behavior. We view this shift from one-shot generation to closed-loop, evidence-centered control as a necessary step for deploying LLM-based text systems responsibly in safety- and operations-critical environments. SRCoT reframes “auditable generation” as a controllable, inspectable feedback process—especially valuable for safety- and operations-critical settings such as emergency management and medical and fire services, where models must often run locally under limited compute and strict local rules, and where human-in-the-loop verification is a non-negotiable safety constraint.

\section{Limitations}
\label{sec:limitations}
Our study has several limitations. \textbf{(1) Evidence scope.} Results are primarily qualitative, targeting auditability, control structure, and gap-driven behavior; they do not establish comprehensive recall, hallucination reduction, clinical utility, or superiority over baselines without larger datasets and controlled evaluations. \textbf{(2) Data scale and retrieval bias.} The EMS scenario is small and potentially selective; with large top-k, BM25 can approach “return everything”, masking precision, context use, and drift, and retrieval remains sensitive to query phrasing and SOP quality. \textbf{(3) System dependence.} Observed follow-up behavior is partly induced by prompts, schemas, and output contracts, reflecting end-to-end design rather than model capability alone. \textbf{(4) Evaluation and simulation limits.} The optional Judge may hallucinate or mis-score evidence, keyword conflict markers are coarse heuristics, and Munich Sim results expose prototype-specific issues (e.g., citation slippage, scope creep, unsafe handling of missing contraindications), limiting transfer across domains, document styles, and languages. \textbf{(5) Deployment framing.} Local execution requires governance for safety, and the cybernetic framing is descriptive of the pipeline, not evidence of biological mechanism replication.

A comprehensive expert-rating study with interrater reliability was not feasible in the present setting because access to a sufficiently large and representative panel of domain experts is constrained by funding, organizational and operational factors, making systematic recruitment and coordination impractical for our study. Additionally, purely subjective ratings we noted early on in our research were vulnerable to presentation effects, since fluent writing style can appear plausible even when claims are incorrect which is a critical issue in safety-critical domains. We therefore adopt a document-anchored validation strategy in which correctness and local compatibility are operationalized via normative references (e.g., Standard Operating Procedures and curated reference lists as \emph{normative baseline} and \emph{binding reference}) and evaluated at the level of individual claims, prioritizing traceability (explicit claim and evidence support) as well as conservative abstention when evidence is insufficient. Local heterogeneity is also treated as a core constraint: SOP availability, age, and bindingness vary across regions and organizations, so evaluations are interpreted relative to the respective local corpus and normative baseline, and cross-region comparisons are limited.

\section{Ethical Concerns}
Both reported case settings use fully synthetic scenarios, without real patient records, radio recordings, or incident data. No human participants were recruited and no clinical interventions were performed for these case studies. The following principles describe the intended research scope and requirements for any future use.

\textbf{Purpose and Research Motivation.}
The framework is designed exclusively as decision support for prehospital and emergency management contexts, with the aim of improving information synthesis, situational awareness, and documentation under time pressure. It is not intended to replace clinical judgment, operational command, or established protocols. Human decision-makers retain full authority and responsibility for all actions taken. The system is framed to encourage verification and reflection rather than automation, explicitly reinforcing human-in-the-loop or human-in-command use as a non-negotiable design constraint.

\textbf{Prevention of Over-Reliance and Accountability.}
To mitigate automation bias, the system is designed to surface uncertainty explicitly and to avoid presenting speculative or inferred details as facts. Structured, fluent outputs should be accompanied by clear indications of missing or uncertain information (e.g., explicit “not specified” markers), and the system is intended to abstain from completing safety-critical fields when inputs are incomplete. Accountability and liability remain with human operators and their governing organizations; ethical deployment therefore requires clear governance structures, documented decision ownership, and policies that define the role of tool outputs as advisory rather than prescriptive.

\textbf{Handling of Harmful or Safety-Critical Content.}
Because unsafe default completions (such as inferred contraindications, timelines, or resource availability) could lead to harm, the framework prioritizes conservative behavior. It is designed to discourage unsupported operational specifics and to defer or flag gaps instead of filling them; the observed failures show that this behavior is not guaranteed. Verification steps are required for critical items, and outputs are intended to prompt human review rather than immediate execution.

\textbf{Data Protection and Confidentiality.}
Although the framework can be executed locally, ethical use still requires rigorous data protection practices. Patient and incident data may persist in prompts, logs, or exported artifacts; therefore, retention limits, access controls, and audit trails are necessary. Ethical deployment assumes compliance with applicable data protection regulations and organizational policies governing sensitive information.

\textbf{Fairness, Context Sensitivity, and Performance Limits.}
Unequal performance across contexts is a recognized risk, as local standard operating procedures and documentation quality vary widely. Ethical use requires validation in the specific operational environments where the system is deployed, ongoing maintenance, and transparent communication of known limitations. The framework should not be assumed to generalize reliably beyond the contexts for which it has been evaluated.

\textbf{Responsible Release, Dual-Use Mitigation, and Policy Compliance.}
The framework has potential dual-use risks, including misuse for autonomous operational control, dispatch decisions, or applications outside approved governance structures. To prevent scope creep, responsible release includes explicit scope restrictions, access controls, usage logging, and policy enforcement mechanisms. The system is not intended for autonomous decision-making or unauthorized operational use, and safeguards are required to ensure it remains within its stated research and support role.

\section{Future Work}
This work presents an early prototype with an exploratory evaluation. Immediate next steps include systematic studies using larger and more diverse local corpora to quantify evidence coverage, failure modes, and constraint violations, as well as controlled ablations (e.g., removing individual components or varying top-$k$) to isolate the contributions of specific design choices. We also plan to investigate stronger retrieval and reranking strategies that remain compatible with edge and offline constraints, alongside formal domain-specific governance mechanisms such as versioned SOP corpora, audit logging, and prompt-injection defenses.

Future work should also broaden the empirical basis of evaluation by generating many more realistic test cases and operational scenarios and reviewing them with domain experts using structured protocols and, where possible, inter-rater agreement.

A second open direction concerns controller and orchestration design, including how many reasoning passes and retrieval--revise loops are required for different task classes, and which prompt elements most effectively elicit metacognitive-like behaviors (e.g., gap detection and conflict checking) that improve outputs within fixed model constraints.

Finally, we will explore mechanisms for state and policy accumulation, including whether modular ``micro-policies'' can be maintained as retrievable rule fragments without destabilizing behavior, and whether writing back compact summaries of prior outputs, verified facts, and open gaps can improve subsequent passes by reducing repetition and sharpening follow-on queries.

A near-term direction is improving the reliability of numerical outputs for aggregation-heavy queries (e.g., totals, deltas, and \emph{Materialverbrauch} as consumption summaries), where evidence is often distributed across many log-like entries (e.g., per-unit ``Materialverbrauch'' lines) and only occasionally summarized in overview/diff blocks. The v1.0 prototype lacks dedicated reconciliation mechanisms and can therefore over-rely on whichever fragments are retrieved, yielding brittle totals or silent inconsistencies (e.g., reporting a fuel total from an overview block while missing additional per-incident log entries). In our ongoing v1.2 prototype we introduce lightweight, offline-compatible improvements that preserve the two-pass SRCoT structure: (i) generic query expansion by mining salient identifiers from the question and first-pass draft (e.g., item IDs such as \texttt{con\_fuel\_mix=500} or incident tags like \texttt{\#000071}) to better steer the second retrieval pass; (ii) context prioritization to surface aggregate/overview evidence earlier in the prompt; and (iii) an automatic numeric conflict detector that triggers a reconcile rewrite when multiple distinct values are observed for the same key. These changes are intended to improve coverage of relevant ``Materialverbrauch'' evidence and the consistency of totals, without adding heavy infrastructure or online dependencies; their effects require controlled evaluation.

More broadly, we view the central research question as interdisciplinary: Under operational decision constraints, when and why do different control, retrieval, and governance strategies succeed or fail, and how can such triggering behavior be validated in practice?

\section*{Declarations}

\subsection*{Use of AI Tools}
AI-assisted tools supported code development, language editing, restructuring, and grammatical refinement. The project documentation identifies DeepL, ChatGPT with GPT-5.2, and Claude Sonnet 4.5; GPT-5.2 assisted development of the Munich-Sim software. OpenAI Codex assisted manuscript revision, consistency and reference checks, and LaTeX preparation on 10 October 2026. Codex also assisted arXiv preparation on 14 October 2026. The vehicle icons shown in Figure 2 were generated using an OpenAI GPT model. These activities are distinct from the local LLM inference examined in the case studies. The authors take responsibility for the study design, implementation, reported observations, interpretation, references, and final text, including AI-assisted content.

\subsection*{Scope and Non-Medical Disclaimer}
This work evaluates the system’s output quality against \emph{project-defined requirements} and provided reference materials, focusing on consistency, traceability, and controllability (e.g., evidence linkage, gap detection, conflict flagging). It is \emph{not} a demonstration of clinical validity, evidence-based medicine (EBM), or guideline adherence, and it does not establish medical correctness or effectiveness of any measures. The system is a general-purpose software prototype and \emph{not} a medical application/device; it is \textit{not} intended for diagnosis, treatment decisions, or emergency management \textit{nor} for usage in automated high-stakes systems without a qualified human in the loop. All examples are illustrative and do \textit{not} replace professional judgment or applicable regulations and validated guidance.

\subsection*{Code Availability}
Prototype source code is available at \url{https://github.com/lowertheentropy/paper-srcot}.

\subsection*{Data Availability}
Partial public exports for selected EMS and Munich-Sim reports are available in the \href{https://github.com/lowertheentropy/paper-srcot/tree/main/artifacts/frontiers-cases}{case artifact archive} of the same repository. They preserve generated answers, drafts, follow-up questions, and Judge outputs without correction. Retrieved contexts and evidence-text payloads are excluded because source provenance and redistribution rights are not yet fully established. A provenance manifest records original filenames, hashes, and verified report/result mappings. These selected exports are not a complete reproduction bundle for all scenarios discussed here. Requests concerning other reports or omitted materials should be directed to the corresponding author.

\subsection*{Related Manuscript}
A companion brief report, \emph{Self-Reflective Chain-of-Thought (SRCoT) for Auditable, Locally SOP-Grounded Text Generation with Edge LLMs in Emergency Management}, has been submitted to Frontiers in Disaster and Emergency Medicine. It shares the architecture and selected EMS and Munich-Sim case material with this extended technical account. The two manuscripts describe overlapping work and must not be interpreted as independent empirical studies.
\subsection*{Funding}
Supported by Open Access funds of Freie Universit\"at Berlin.

\subsection*{Competing Interests}
The authors declare that they have no competing interests.

\subsection*{Author Contributions}
Fabian Broszio conducted the study and wrote the original draft. Dr. Michael Hoffmann reviewed and edited the text and contributed anthropological expertise. Prof. Dr. Adrian Paschke provided supervision and reviewed the manuscript. All authors approved the final version.

\bibliography{sn-bibliography}

\end{document}